\documentclass[review]{fcs}
\usepackage{url}            
\usepackage{graphicx} 
\usepackage{subfigure}
\usepackage{wrapfig}
\usepackage{algorithm}
\usepackage{algorithmic}

\usepackage{bbm}
\usepackage{bm}
\usepackage{amsmath}
\usepackage{amssymb}
\usepackage{amsthm}
\usepackage{mathrsfs}
\usepackage{color}
\usepackage{graphicx}
\usepackage{algorithm}
\usepackage{algorithmic}
\usepackage{subfigure}
\usepackage{epsfig}
\usepackage{epsf}
\usepackage{booktabs}
\usepackage{tabularx}
\usepackage{tabulary}
\usepackage{multirow}
\usepackage{graphics}
\usepackage{longtable}
\usepackage{subfigure}  
\usepackage[numbers]{natbib}

\usepackage{hyperref}

\usepackage{bm}
\usepackage{amsmath}
\usepackage{verbatim}
\usepackage{bm}

\usepackage{url}           
\usepackage{graphicx} 
\usepackage{subfigure}

\def\c{{\bf c}}

\def\g{{\bf g}}

\def\l{{\bf l}}

\def\x{{\bf x}}
\def\y{{\bf y}}

\def\XX{{\mathcal X}}
\def\YY{{\mathcal Y}}

\def\bbE{{\mathbb E}}

\def\bbR{{\mathbb R}}

\title{Robust Semi-Supervised Learning in Open Environments}
\author[1,2]{Lan-Zhe Guo}
\author[1]{Lin-Han Jia}
\author[1]{Jie-Jing Shao}
\author*[1,3]{Yu-Feng Li}
\address[1]{National Key Laboratory for Novel Software Technology, Nanjing
University, China}
\address[2]{School of Intelligence Science and Technology, Nanjing University, China}
\address[3]{School of Artificial Intelligence, Nanjing University, China}

\fcssetup{
  received       = {month dd, yyyy},
  accepted       = {month dd, yyyy},
  corr-email     = {xxx@xxx.xxx},
}

\begin{abstract}
Semi-supervised learning (SSL) aims to improve performance by exploiting unlabeled data when labels are scarce. Conventional SSL studies typically assume close environments where important factors (e.g., label, feature, distribution) between labeled and unlabeled data are consistent. However, more practical tasks involve open environments where important factors between labeled and unlabeled data are inconsistent. It has been reported that exploiting inconsistent unlabeled data causes severe performance degradation, even worse than the simple supervised learning baseline. Manually verifying the quality of unlabeled data is not desirable, therefore, it is important to study robust SSL with inconsistent unlabeled data in open environments. This paper briefly introduces some advances in this line of research, focusing on techniques concerning label, feature, and data distribution inconsistency in SSL, and presents the evaluation benchmarks. Open research problems are also discussed for reference purposes.
\end{abstract}
\begin{document}

\section{Introduction}
Semi-supervised learning (SSL) is an effective learning paradigm to improve learning performance by attempting to exploit abundant unlabeled data when labels are scarce. It has been reported that, in certain cases, such as image classification, SSL methods can achieve the performance of purely supervised learning even when a substantial portion of the labels in a given data set has been discarded~\cite{wang2022usb}.

\begin{figure*}[t]
	\centering
	\includegraphics[width=\linewidth]{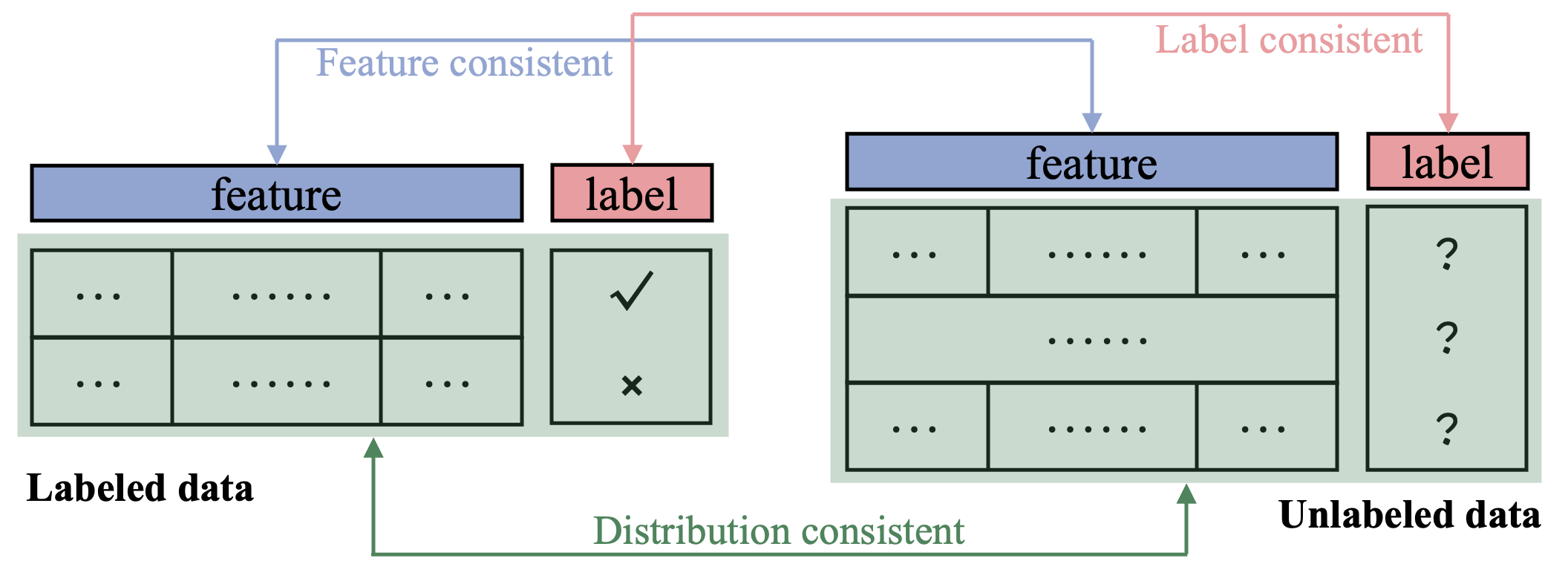}
	\caption{Typical consistent factors assumed in close environment semi-supervised learning.}
	\label{fig:ssl}
\end{figure*}

It is noticeable that the current success of SSL is mostly based on the close environment assumption where important factors between the labeled and unlabeled data are consistent. For example, all the unlabeled instances should belong to the class label set in labeled data, the features describing labeled and unlabeled data should be the same, and all labeled and unlabeled data should be sampled from an identical distribution. Figure~\ref{fig:ssl} illustrates those consistent factors assumed in close environment SSL studies.

However, many real-world 
applications involve open environments~\cite{zhou2022open} where class label set, feature space, and data distribution could be inconsistent between labeled and unlabeled data. The main reason lies in the fact that the collection process of unlabeled data is different from labeled data, which lacks human supervision and easily collects data that are inconsistent with the target task. It is also impossible to manually validate the quality of unlabeled data, otherwise it goes against the SSL’s purpose of reducing human labor. It has been widely reported that SSL suffers severe performance degradation problems with inconsistent unlabeled data and could be even worse than the simple supervised learning baseline which does not exploit more unlabeled data~\cite{guo2018general, oliver18robust, guo2020safe, li2019towards, safessl,dualmatch}. Such phenomena undoubtedly violate the expectations of SSL and limit its effectiveness in more practical tasks.

It seems the robust SSL in open environments is relevant to studies like out-of-distribution (OOD) detection~\cite{ood,zhou2023ods,decoop,shao2022log}, or open set recognition~\cite{DBLP:journals/pami/GengHC21,DBLP:journals/ml/ShaoYG24}. However, these studies either assume that there is an accurate classification model or sufficient labeled data, which limits their application in SSL. There are also some unsupervised OOD detection studies~\cite{SSD} utilizing the power of the contrastive learning~\cite{contrastive} framework to learn better representation for OOD detection. However, although these studies do not require labels, they still need a large amount of in-distribution data for training, while in open environment SSL it is difficult to get the clean in-distribution unlabeled data.

Despite the grand challenges, many research efforts have recently been devoted to robust SSL in open environments. This paper will briefly introduce some advances in this line of research, focusing on approaches concerning inconsistent labels, inconsistent features, and inconsistent distributions between labeled and unlabeled data. Moreover, we introduce the benchmark dataset and performance measures applicable to evaluate the robustness of SSL in open environments and provide a public SSL toolkit for related research. Open research problems are also discussed for reference purposes.

\section{Robust SSL in Open Environments}

In the SSL task, we are given a set of training data which includes labeled data set $\mathcal{D}_{l}$ consists of $n$ labeled instances $\mathcal{D}_{l} = \{(\x_1, \y_1), \cdots, (\x_n, \y_n)\}$ and unlabeled data set $\mathcal{D}_{u}$ consists of $m$ unlabeled instances $\mathcal{D}_{u} = \{\x_{n+1}, \cdots, \x_{n+m}\}$. Usually, $m \gg n$, $\x \in \XX \in \bbR^d$, $\y \in \YY=\{1,\cdots,K\}$ where $d$ is the number of feature dimension and $K$ is the number of classes. The goal of SSL is to learn a model $f(\x;\theta): \{\XX; \Theta\} \to \YY$ parameterized by $\theta \in \Theta$ from training data to minimize the generalization risk $R(f)=\bbE_{(X,Y)}[\ell(f(X;\theta),Y)]$, where $\ell:\YY \times \YY \to \bbR$ refers to certain loss function, e.g., mean squared error or cross-entropy loss.

In open environments, unlabeled data could be inconsistent with labeled data in terms of class label space, feature dimension, and data distribution. We denote the degree of inconsistency as $t\in[0,1]$. A higher $t$ indicates a greater inconsistency, i.e., more unlabeled instances that are inconsistent with the target task. The robust SSL studies in open environments aim to decrease the negative impact of inconsistent unlabeled data, on the one hand, improve the performance via exploiting unlabeled data, and on the other hand, in the worst case, the SSL performance should not be worse than the supervised learning baseline which does not exploit more unlabeled data.

\section{Label Inconsistent}

\begin{figure}[t]
	\centering
	\includegraphics[width=\linewidth]{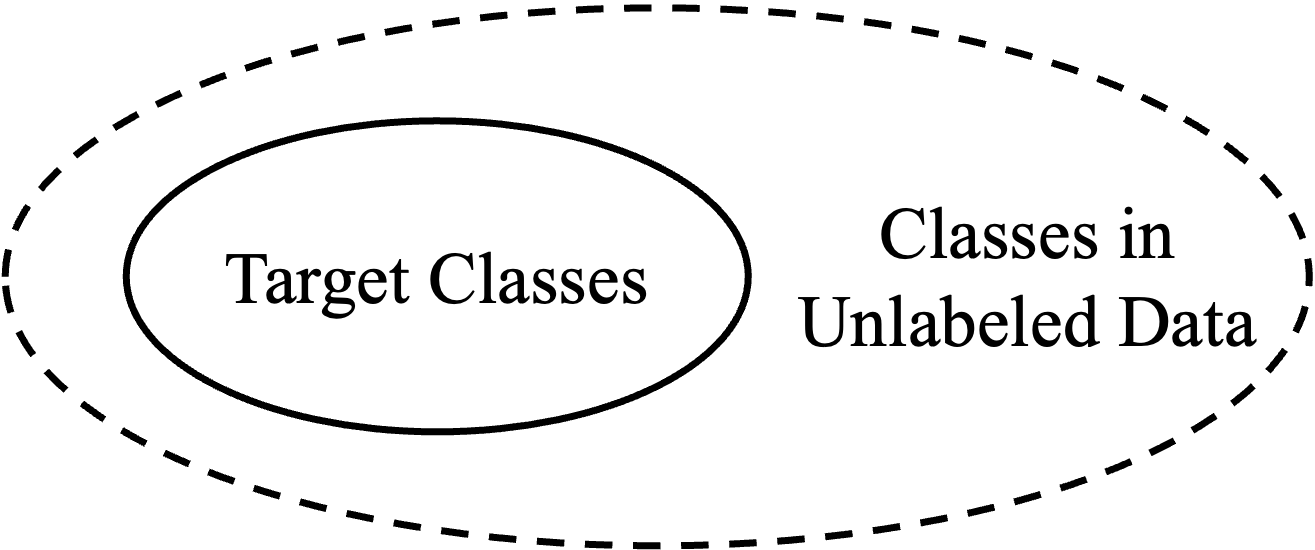}
	\caption{Irrelevant classes occur in the unlabeled dataset.}
	\label{fig:classes}
\end{figure}

Close-environment SSL studies typically assume that the class label of any unlabeled instances should be a member of the given label space $\YY$. However, this assumption does not always hold. This is because unlabeled data is much easier to collect than labeled data in real-world applications, and the collection process of unlabeled data has less human verification. Thus, it is more likely for unlabeled data to have unseen classes that are irrelevant to the target task. For example, in the image classification task, unlabeled images crawled from Internet/social networking according to keywords usually contain broader category concepts than labeled data~\cite{yang2011can,guo2020safe,li2019towards,jia2024realistic}. We illustrate label inconsistency in Figure~\ref{fig:classes} to help understand the problem.

Many researchers have pointed out that SSL is not robust to irrelevant unseen classes of unlabeled data, and could perform even worse than the simple supervised learning method that uses only a small number of labeled data~\cite{oliver18robust, guo2020safe}. 

The straightforward idea to deal with this problem is to detect and remove these irrelevant unseen class unlabeled instances. It is noteworthy that this problem is different from OOD detection since OOD detection approaches typically require a large corpus of in-distribution labeled data and would fail due to the scarcity of labeled data in SSL. Recently, a simple yet effective approach was proposed to tackle the semi-supervised OOD detection issue~\cite{ZhouGCLP21}. This approach learns a new representation space via a novel distance measure in which OOD samples could be separated well with limited labeled data and in-distribution data.

Some approaches try to decrease the negative impact of these unseen class unlabeled instances in the training process. Various scoring mechanisms have been proposed to evaluate how much contribution an unlabeled instance has to the model training~\cite{chen20,yu2020multi,saito2021openmatch}. If the score is higher than the threshold the instance is retained, otherwise, the instance is discarded. In addition to the hard threshold, some works try to assign soft weight to the unlabeled training instances~\cite{peng2019investigating,guo2020safe}.

Instead of treating all irrelevant classes unlabeled instances as harmful, some researchers find these instances could also be helpful for model training. One promising way is to exploit the irrelevant unlabeled instances to help learn better representations via the self-supervised learning paradigm~\cite{huang2021trash}.

Robust SSL focuses on how to avoid the negative impact of unseen class unlabeled data. Meanwhile, some studies focus on the setting that the unseen classes in unlabeled data also need to be classified, which is called open-world SSL~\cite{ow_ssl,guo2022NACH}. This line of research is more like class-incremental learning~\cite{class-incremental}, and different from the goal of robust SSL in open environments.

\section{Feature Inconsistent}

\begin{figure}[t]
	\centering
	\includegraphics[width=\linewidth]{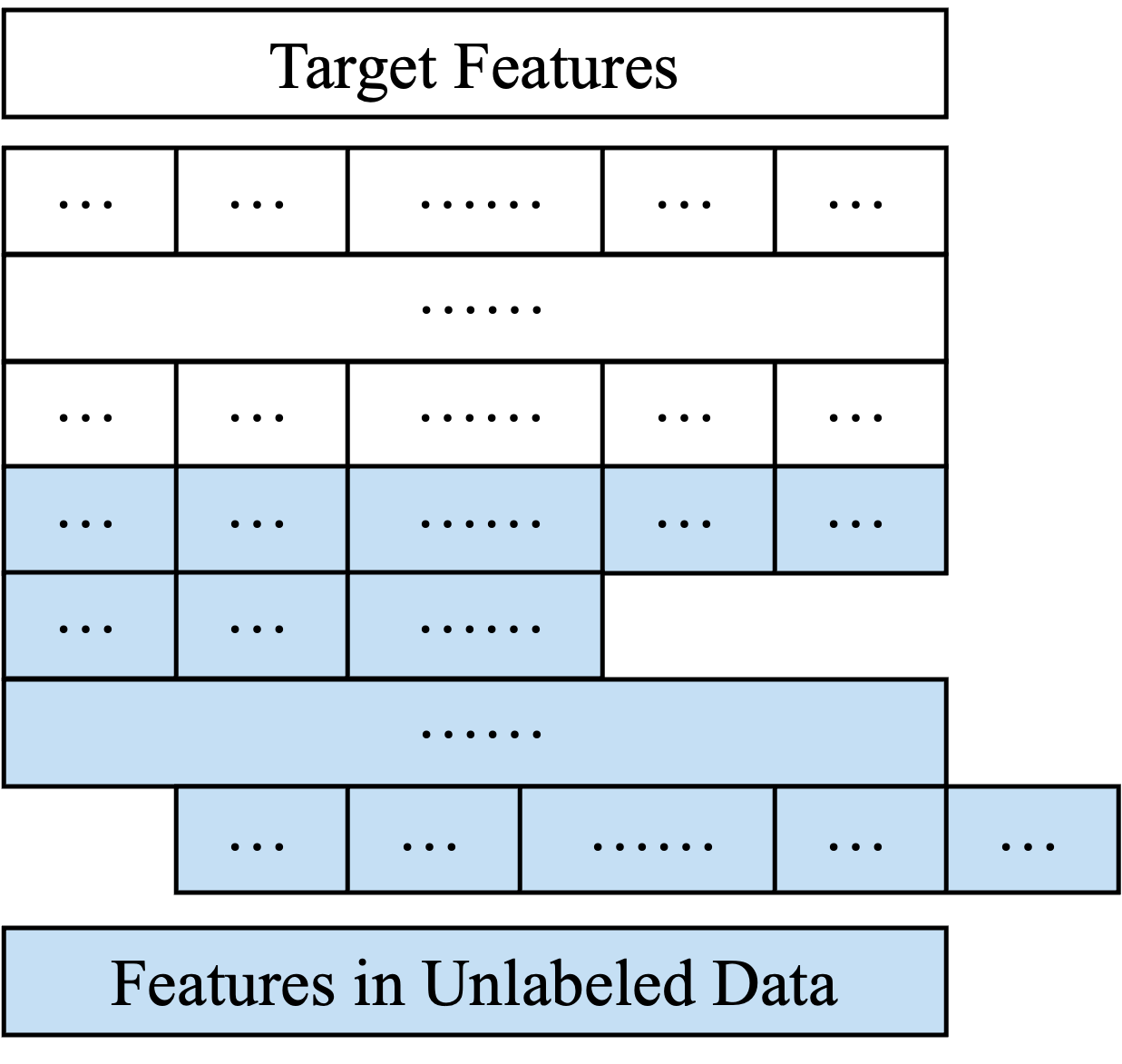}
	\caption{Feature inconsistent in SSL.}
	\label{fig:feature}
\end{figure}

Close-environment SSL studies typically assume that all unlabeled instances reside in the same feature space with the labeled data. Unfortunately, this does not always hold. For example, in the image classification task, the labeled data are all color images while the unlabeled data could contain grayscale images, resulting in the loss of two color channels. In tasks dealing with tabular data, such as financial analysis tasks or recommendation systems~\cite{tabular}, decremental or incremental features in unlabeled data are more common. Figure~\ref{fig:feature} illustrates the feature inconsistent problem in SSL.

As pointed out by~\cite{jia2024realistic}, close environment SSL methods could suffer severe performance degradation when facing the feature inconsistent between labeled and unlabeled data.

Compared with the label inconsistent problem, detecting which unlabeled instances have inconsistent features with the target task is much easier since validating the feature $\x$ is irrelevant to the label. Therefore, the straightforward method to address the feature inconsistent problem is to detect and remove all inconsistent unlabeled instances. 

However, this baseline method cannot effectively utilize the information behind these unlabeled data, resulting in limited performance improvement. Another approach that readily comes to mind is to remove all incremental features and fill in decremental features, whereas how to fill the missing feature to ensure the SSL performance will not degrade is still a challenging problem~\cite{jia2024realistic}.

There are also studies focusing on the adversarial feature inconsistency unlabeled examples, which can be categorized into two aspects: \emph{Attack} and \emph{Defense}. Semi-supervised attack techniques study how to generate adversarial unlabeled examples that cause SSL predictions to be incorrect. The major techniques can be categorized as misleading sequence injection, which aims to inject a sequence
of synthetically unlabeled examples into the unlabeled training data and cause the model to make wrong predictions~\cite{Carlini21} and adversarial perturbation generation, which aims to learn a feature perturbation generator for the training examples and making the model output wrong predictions when training with these perturbed examples~\cite{YanLTWLCP21,LiuSZ0H19}. The defense techniques study how to make the SSL robust to adversarial unlabeled examples. The major techniques can be categorized as robust regularization, which aims to design regularization terms to the objective of SSL directly~\cite{MiyatoMKI19, YuWMZ19} and the combination with classical distribution robust optimization methods~\cite{NajafiMKM19}.

Recent studies of robust SSL with feature inconsistency mainly focus on the image classification task. It is noteworthy that tabular data is also commonly encountered in real scenarios~\cite{tabular}. Compared with image data, the feature inconsistent problem is more commonly occurring in tabular data. Robust SSL with inconsistent features on tabular data is an important yet understudied problem.

\section{Distribution Inconsistent}

\begin{figure}[t]
	\centering
	\includegraphics[width=\linewidth]{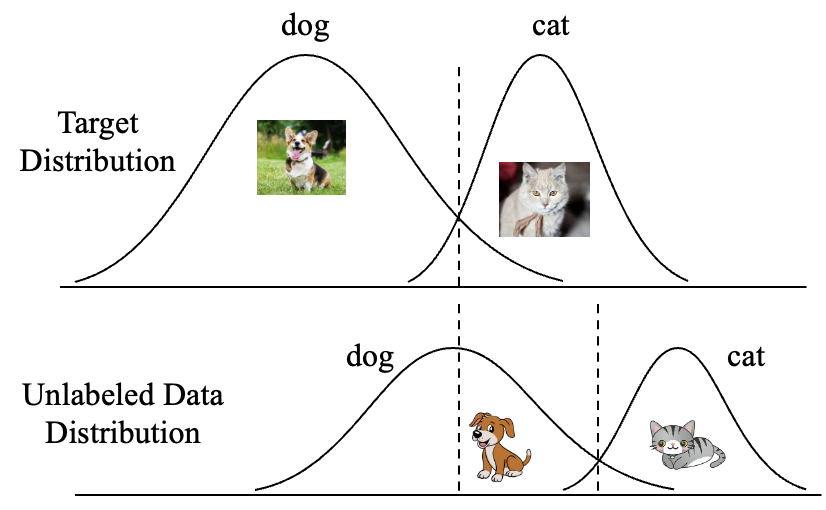}
	\caption{Distribution inconsistent between labeled and unlabeled data in SSL.}
	\label{fig:distribution}
\end{figure}

Close-environment SSL studies typically assume that all labeled and unlabeled data are independent samples from an identical distribution (i.e., $i.i.d.$ samples). Unfortunately, this does not always hold. Taking the image classification as an example again, the labeled data may sampled from natural images. In contrast, the unlabeled data may be selected from the internet according to some keywords and may include cartoon images~\cite{Sword++}. These problems also commonly happen in scenarios like sentiment analysis~\cite{guo2020safe}, remote sensing~\cite{CLIP-SSL}, legal judgment~\cite{lawgpt}, etc. Figure~\ref{fig:distribution} illustrates that ignoring the data distribution inconsistent mismatch may lead to seriously downgraded performance.

There have been plentiful studies concerning distribution shifts such as prior probability shifts, covariate shifts, and concept shifts. However, the relevant studies mainly focus on the training/testing distribution change and are conducted under the umbrella of domain adaptation or transfer learning~\cite{pan2009survey}. In SSL studies the distribution occurs within the training data. To be able to handle various kinds of data distribution inconsistent between labeled and unlabeled data is an important requirement for robust SSL in open environments. 

The straightforward method is to treat labeled data as the target domain and unlabeled data as the source domain and then apply domain adaptation techniques to learn new representations for all training instances to eliminate the distribution mismatch~\cite{chen2019distributionally,huang2021universal,guo2020record}. However, due to the label scarcity in SSL, these methods can only consider the adaptation in an unsupervised manner and ignore task-related label information.

Recent work presented a theoretical framework that presents three main reasons why SSL algorithms can not perform well with inconsistent distributions: coupling between the pseudo-label predictor and the target predictor, biased pseudo labels, and restricted instance weights~\cite{jia2023bidirectional}. To address these challenges, they provided a new method called bidirectional adaptation that can adapt to the distribution of unlabeled data for debiased pseudo-label prediction and to the target distribution for debiased target prediction, thereby mitigating the above shortcomings.

Moreover, some works focus on the problem that unlabeled data distribution is long-tailed and report that SSL suffers performance degradation on tail classes~\cite{kim2020distribution,wei2021crest, LIMI, guo2022class,wei2024transfer}. Conventional long-tailed techniques can not be applied directly due to the label scarcity in SSL. The general principle is to design distribution alignment techniques to calibrate the distribution of pseudo-labels to align with the target distribution~\cite{kim2020distribution, wei2021crest,guo2022class}. 

\section{Evaluation Benchmark}
Conventional SSL studies mainly evaluate performance on standard image classification datasets and report classification accuracy. How to fair evaluate the robustness of SSL methods in open environments is under-considered. In this section, we briefly introduce some datasets, applicable performance measures, and an open-sourced SSL toolkit.

\subsection{Datasets}
Constructing open environment SSL benchmarks that contain different extents of inconsistency between labeled and unlabeled data is important for the evaluation of robust SSL algorithms. Recently, a more realistic SSL benchmark included both label, feature, and distribution inconsistent has been provided~\cite{jia2024realistic}. The benchmark involves various data types in machine learning, including tabular datasets from the UCI dataset, widely applied image datasets, and text datasets. Specifically, to simulate the inconsistent labels, they construct inconsistent labeled space by randomly selecting some classes and discarding the labeled data belonging to these classes. To simulate the inconsistent features, they randomly mask features for tabular data and convert the images to grayscale, resulting in the loss of two color channels for image datasets. For the text datasets, they employ text truncation, and truncated portions are filled with ``$<pad>$". To simulate the data distribution, for image and text datasets, they adopt the Image-CLEF~\cite{imageclef} and the IMDA/Amazon~\cite{mcauley2013hidden,maas2011learning} to construct the labeled and unlabeled data which are natural distribution shifts. For tabular datasets, they calculate the centroids of each class and use the distance between instances and class centroids to filter instances, thus constructing an environment with inconsistent data distribution.

\subsection{Performance Measures}
To achieve a fair and comprehensive evaluation of robust SSL in open environments, only reporting the classification accuracy or error is not enough. A series of performance metrics tailored for robust SSL in open environments have been proposed recently~\cite{jia2024realistic}. These metrics begin by defining a function $Acc(t)$, which quantifies the change in classification accuracy as a function of the inconsistency level $t$. This function is used to construct the Robustness Analysis Curve (RAC) that maps the inconsistency level $t$ to the corresponding accuracy $Acc(t)$. Unlike conventional SSL evaluations that focus solely on $Acc(0)$ or a specific  $Acc(t)$, various performance metrics are proposed based on the RAC that include Area Under the RAC Curve (AUC) which captures the overall robustness of SSL approaches; Expected Accuracy (EA) which describes the average performance across all inconsistency levels; Worst-Case Accuracy (WA) which identifies the lowest accuracy level, representing performance in the worst-case scenario; Expected Variation Magnitude (EVM) which captures the average magnitude of performance variation; Variation Stability (VS) which quantifies the stability of the performance variation; Robust Correlation Coefficient (RCC) which captures the overall trend of performance variation. The detailed formulation of these metrics is presented in Table~\ref{metrics}.

\begin{table*}[t]
    \centering
    \caption{Performance metrics for robust SSL in open environments. $Acc(t)$ describe the change in classification accuracy with the inconsistency extent $t$, $P_T(t)$ is the distribution for $t$, $Acc'(\cdot)$ indicate the first derivative.}
    \label{metrics}
    \begin{tabular}{l|c}
    \hline \hline
     Area Under the Curve (AUC)    &  $\text{AUC}(Acc) =\int_0^1 Acc(t)dt$ \\ 
     \hline 
     Expected Accuracy (EA) &  $\text{EA}(P_T,Acc) = \langle P_T,Acc\rangle=\int_0^1 P_T(t)Acc(t)dt$ \\
     \hline 
     Worst-Case Accuracy (WA) & $\text{WA}(Acc)=\min_{t\in[0,1]} Acc(t)$ \\ 
     \hline 
     Expected Variation Magnitude (EVM) & $\text{EVM}(Acc) =\int_0^1 |Acc'(t)|dt$ \\
     \hline
     Variation Stability (VS) & $\text{VS}(Acc)=\int_0^1 [Acc'(t)-(\int_0^1Acc'(t)dt)]^2dt$ \\
     \hline
     Robust Correlation Coefficient (RCC) & $\text{RCC}(Acc)=\frac{\int_0^1 Acc(t)\cdot t dt - \int_0^1 Acc(t) dt}{\sqrt{\int_0^1 t^2dt -1}\cdot\sqrt{\int_0^1 Acc^2(t)dt-(\int_0^1 Acc(t) dt)^2}}$ \\
     \hline\hline
    \end{tabular}
\end{table*}

\subsection{Open-Sourced Toolkit}
To provide easier evaluation and implementation of SSL algorithms, an open-sourced SSL toolkit: LAMDA-SSL is released~\cite{lamdassl}. LAMDA-SSL incorporates more than 30 SSL algorithms, supports various data types, and is compatible with other popular machine learning toolkits such as ``scikit-learn" and ``pytorch". The toolkit is available at \url{https://ygzwqzd.github.io/LAMDA-SSL}.

\section{Open Challenges}
Though robust SSL in open environments has attracted much attention, it is still in its infancy. We hope to propose new research directions to broaden and boost robust SSL research.

\textbf{Theoretical Issues}. Many theoretical problems about robust SSL have not been addressed yet. For example, when the inconsistent unlabeled data is helpful or harmful, how the generalization performance varies with different inconsistent extents, etc. More efforts are desired to be devoted.

\textbf{General Data Types}. SSL studies mainly focus on homogeneous data, especially image data. Tabular data is also a commonly occurring data in practical tasks~\cite{tabular, ye2024closer}. The heterogeneous property of tabular data causes the failure of SSL algorithms. For example, consistency regularization, which is the most important technique in SSL, encourages the model to have similar output distribution on an instance and its augmented variants. The notion of augmentation simply does not exist in tabular data. Therefore, there is an urgent need to develop robust SSL techniques for more general data types.

\textbf{Exploiting Pre-Trained Models}. With the success of the ``pre-train and fine-tune" paradigm, more and more pre-trained models have been released. Similar to the goal of SSL, selecting and adapting helpful pre-trained models can also decrease the labeled data requirement for the target task~\cite{guo2023identifying}. Thus, how to bridge the pre-trained model with SSL is a promising direction.  Recently, there have been some studies that have tried to exploit SSL techniques with large language models~\cite{LLM_SSL} and vision-language models~\cite{CLIP-SSL}. However, the robustness of these methods after exploiting more unlabeled data is still an unaddressed problem.

\textbf{From Perception to Decision-making}. 
Current SSL studies mainly focus on perceptual tasks such as image classification, while practical tasks often encounter decision-making tasks that involve interaction with the environment. The dynamic of environments poses significant challenges to robustness, meanwhile, high-quality data is expensive in decision-making tasks, posing a great need for SSL. Many researchers have been exploring how to utilize unlabeled data for reinforcement learning~\cite{DBLP:conf/icml/YuKCHFL22,KDD-offline} on these tasks, including reward-free or action-free data~\cite{zheng2023semi,li2023imitation}. Therefore, it is important to broaden robust SSL studies into decision-making tasks with interactive environments. 

\section{Conclusions}
This paper introduces open environments SSL. We present a definition of this problem, in which unlabeled data could contain label/feature/distribution inconsistent with the target task, and briefly introduce some research advances in this line of research.  Although we consider these inconsistent problems separately, in practice they often occur simultaneously. It can hardly be a thorough review of all the relevant work and is mostly a brief review of general principles and strategies rather than specific learning algorithms. The quality of unlabeled data is hard to validate and it is fundamentally important to enable SSL to achieve excellent performance in the usual case while keeping satisfactory performance no matter what unexpected unfortunate issues occur in unlabeled data. This is crucial for achieving robust SSL in practical tasks.

\begin{acknowledgement}
This research was supported by the National Science Foundation of China (62306133, 62176118).
\end{acknowledgement}

\bibliographystyle{fcs}
\bibliography{paper}


\end{document}